\pdfoutput=1
\documentclass[runningheads]{llncs}
\usepackage{graphicx}
\usepackage{amsmath,amssymb} % define this before the line numbering.
\usepackage{color}
\usepackage{algorithm}
\usepackage{algorithmic}
\graphicspath{{./figs/}}
\newcommand*{\eg}{e.g. }

\begin{document}
\title{Dist-GAN: An Improved GAN using Distance Constraints} 
% Replace with your title

\titlerunning{Dist-GAN: An Improved GAN using Distance Constraints}
% Replace with a meaningful short version of your title
%
\author{Ngoc-Trung Tran \orcidID{0000-0002-1308-9142} \and
Tuan-Anh Bui \orcidID{0000-0003-4123-2628} \and
Ngai-Man Cheung \orcidID{0000-0003-0135-3791}}
%
%Please write out author names in full in the paper, i.e. full given and family names. 
%If any authors have names that can be parsed into FirstName LastName in multiple ways, please include the correct parsing, in a comment to the volume editors:
%\index{Lastnames, Firstnames}
%(Do not uncomment it, because you may introduce extra index items if you do that, we will use scripts for introducing index entries...)
\authorrunning{Ngoc-Trung Tran, Tuan-Anh Bui and Ngai-Man Cheung}
% Replace with shorter version of the author list. If there are more authors than fits a line, please use A. Author et al.
%

\institute{ST Electronics - SUTD Cyber Security Laboratory\\
	Singapore University of Technology and Design\\
\email{\{ngoctrung\_tran,tuananh\_bui,ngaiman\_cheung\}@sutd.edu.sg}}
\maketitle              % typeset the header of the contribution

\begin{abstract}

We introduce effective training algorithms for  Generative Adversarial Networks (GAN)  to alleviate mode collapse and gradient vanishing. In our system,  we constrain the generator by an  Autoencoder (AE). We propose a formulation to consider the reconstructed samples from AE as ``real'' samples for the discriminator. This couples the convergence of the AE with that of the discriminator, effectively slowing down the convergence of discriminator and reducing gradient vanishing.
Importantly, we propose two novel distance constraints to improve the generator. First, we propose a {\em latent-data distance constraint} to enforce compatibility between the latent sample distances and the corresponding data sample distances. We use this constraint to explicitly prevent the generator from mode collapse. Second, we propose a {\em discriminator-score distance constraint} to align the distribution of the generated samples with that of the real samples through the discriminator score. We use this constraint to guide the generator to synthesize samples that resemble the real ones. Our proposed GAN using these {\bf distance} constraints, namely {\bf Dist}-GAN, can achieve better results than state-of-the-art methods across benchmark datasets: synthetic, MNIST, MNIST-1K, CelebA, CIFAR-10 and STL-10 datasets. Our code is published here\footnote{https://github.com/tntrung/gan} for research.


\end{abstract}

\section{Introduction}
Generative Adversarial Network \cite{goodfellow-nisp-2014} (GAN) has become a dominant approach for learning generative models. It can produce very visually appealing samples with few assumptions about the model. GAN can produce samples \textit{without} explicitly estimating data distribution, \eg in analytical forms. GAN has two main components which compete  against each other, and they improve through the competition. The first component is the generator $G$, which takes low-dimensional random noise $\mathrm{z} \sim P_\mathrm{z}$ as an input and maps them into high-dimensional data samples,  $\mathrm{x} \sim P_\mathrm{x}$. The prior distribution $P_\mathrm{z}$ is often uniform or normal. Simultaneously, GAN uses the second component,  a discriminator $D$, to distinguish whether samples are drawn from the generator distribution $P_G$ or data distribution $P_\mathrm{x}$. Training GAN is an adversarial process: while the discriminator $D$ learns to better distinguish the real or fake samples, the generator $G$ learns to confuse the discriminator $D$ into accepting its outputs as being real. The generator $G$ uses discriminator's scores as feedback to improve itself over time, and eventually can approximate the data distribution.

Despite the encouraging results, GAN is known to be hard to train and requires careful designs of model architectures \cite{goodfellow-nips-2016,radford-arxiv-2015}. For example, the imbalance between discriminator and generator capacities often leads to convergence issues, such as gradient vanishing and mode collapse. Gradient vanishing occurs when the gradient of discriminator is saturated, and the generator has no informative gradient to learn. It occurs when the discriminator can distinguish very well between ``real" and ``fake" samples, before the generator can approximate the data distribution. Mode collapse is another crucial issue. In mode collapse, the generator is collapsed into a typical parameter setting that it always generates small diversity of samples. 

Several GAN variants have been  proposed \cite{radford-arxiv-2015,metz-arxiv-2016,salimans-nisp-2016,arjovsky-arxiv-2017,yazici-arxiv-2018} to solve these problems. Some of them are Autoencoders (AE) based GAN. AE explicitly encodes data samples into latent space and this allows representing data samples with lower dimensionality. It not only has the potential for stabilizing GAN but is also applicable for other applications, such as dimensionality reduction. AE was also used as part of a prominent class of generative models, Variational Autoencoders (VAE) \cite{kingma-arxiv-2013,rezende-icml-2014,burda-arxiv-2015}, which are attractive for learning inference/generative models that lead to better log-likelihoods \cite{wu-arxiv-2016}. These encouraged many recent works following this direction. They applied either encoders/decoders as an inference model to improve GAN training \cite{dumoulin-arxiv-2016,donahue-arxiv-2016,li-arxiv-2017}, or used AE to define the discriminator objectives \cite{zhao-arxiv-2016,berthelot-arxiv-2017} or generator objectives \cite{che-arxiv-2016,warde-arxiv-2016}. Others have proposed to combine AE and GAN \cite{makhzani-arxiv-2015,larsen-arxiv-2015}. 

In this work, we propose a new design to unify AE and GAN. Our design can  stabilize  GAN training, alleviate the gradient vanishing and mode collapse issues, and better approximate data distribution. Our main contributions are two novel distance constraints to improve the generator. First, we propose a {\em latent-data distance constraint}. This enforces compatibility between latent sample distances and the corresponding data sample distances, and as a result, prevents the generator from producing many data samples that are close to each other, i.e. mode collapse. Second, we propose a {\em discriminator-score distance constraint}. This aligns the distribution of the fake samples with that of the real samples and  guides the generator to synthesize samples that resemble the real ones. We propose a novel formulation to align the distributions through the discriminator score. Comparing to state of the art methods using synthetic and benchmark datasets, our method achieves better stability, balance, and competitive standard scores.

\section{Related Works}

The issue of non-convergence  remains an important problem for GAN research, and  gradient vanishing and mode collapse are the most important problems \cite{goodfellow-nips-2016,arjovsky-arxiv-2017a}. Many important variants of GAN have been proposed to tackle these issues. Improved GAN \cite{salimans-nisp-2016} introduced several techniques, such as feature matching, mini-batch discrimination, and historical averaging, which drastically reduced the mode collapse. Unrolled GAN \cite{metz-arxiv-2016} tried to change optimization process to address the convergence and mode collapse. \cite{arjovsky-arxiv-2017} analyzed the convergence properties for GAN. Their proposed GAN variant, WGAN, leveraged the Wasserstein distance and demonstrated its better convergence than Jensen Shannon (JS) divergence, which was used previously in vanilla GAN \cite{goodfellow-nisp-2014}. However, WGAN required that the discriminator must lie on the space of 1-Lipschitz functions, therefore, it had to enforce norm critics to the discriminator by weight-clipping tricks. WGAN-GP \cite{gulrajani-arxiv-2017} stabilized WGAN by alternating the weight-clipping by penalizing the gradient norm of the interpolated samples. Recent work SN-GAN \cite{miyato-iclr-2018} proposed a weight normalization technique, named as spectral normalization, to slow down the convergence of the discriminator. This method controls the Lipschitz constant by normalizing the spectral norm of the weight matrices of network layers.

Other work has integrated AE into the GAN. AAE \cite{makhzani-arxiv-2015} learned the inference by AE and matched the encoded latent distribution to given prior distribution by the minimax game between encoder and discriminator. Regularizing the generator with AE loss may cause the blurry issue. This regularization can not assure that the generator is able to approximate well data distribution and overcome the mode missing. VAE/GAN \cite{larsen-arxiv-2015} combined VAE and GAN into one single model and used feature-wise distance for the reconstruction. Due to depending on VAE \cite{kingma-arxiv-2013}, VAEGAN also required re-parameterization tricks for back-propagation or required access to an exact functional form of prior distribution. InfoGAN \cite{chen-arxiv-2016} learned the disentangled representation by maximizing the mutual information for inducing latent codes. EBGAN \cite{zhao-arxiv-2016} introduced the energy-based model, in which the discriminator is considered as energy function minimized via reconstruction errors. BEGAN \cite{berthelot-arxiv-2017} extended EBGAN by optimizing Wasserstein distance between AE loss distributions. ALI \cite{dumoulin-arxiv-2016} and BiGAN \cite{donahue-arxiv-2016} encoded the data into latent and trained jointly the data/latent samples in GAN framework. This model can learn implicitly encoder/decoder models after training. MDGAN \cite{che-arxiv-2016} required two discriminators for two separate steps: manifold and diffusion. The manifold step tended to learn a good AE, and the diffusion objective is similar to the original GAN objective, except that the constructed samples are used instead of real samples.

In the literature, VAEGAN and MDGAN are most related to our work in term of using AE to improve the generator. However, our design is  remarkably different: (1) VAEGAN combined KL divergence and reconstruction loss to train the inference model. With this design, it required an exact form of prior distribution and re-parameterization tricks for solving the optimization via back-propagation. In contrast, our method constrains AE by the data and latent sample distances. Our method is applicable to any prior distribution. (2) Unlike MDGAN, our design  does not require two discriminators.  (3) VAEGAN considered the reconstructed samples as ``fake", and MDGAN adopts this similarly in its manifold step. In contrast, we use them as ``real" samples, which is important to restrain the discriminator in order to avoid gradient vanishing, therefore, reduce mode collapse. (4) Two of these methods regularize G simply by  reconstruction loss. This is inadequate to  solve the mode collapse. We conduct an  analysis  and explain why additional regularization is needed for AE. 
Experiment results demonstrate that our model outperforms MDGAN and VAEGAN.

\section{Proposed method}
Mode collapse is an important issue for GAN. In this section, we first propose  a new way to visualize the mode collapse. Based on the visualization results, we propose a new model, namely Dist-GAN, to solve this problem.
\subsection{Visualize mode collapse in latent space}
\begin{figure}[t]
\centering
\includegraphics[scale=0.38]{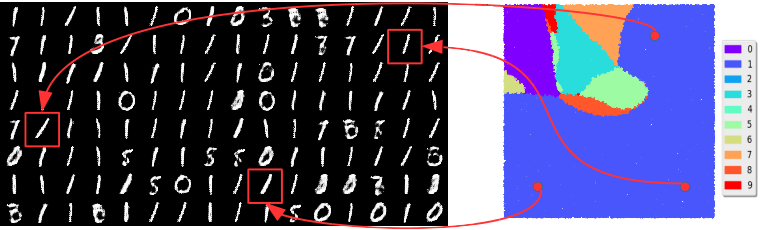}
\caption{(a) Mode collapse observed by data samples of the MNIST dataset, and (b) their corresponding latent samples of an uniform distribution. Mode collapse occurs frequently when the capacity of networks is small or the design of generator/discriminator networks is unbalance.}
\label{collapsed_mnist_dcgan}
\end{figure}
\begin{figure*}[t]
\centering
\includegraphics[scale=0.11]{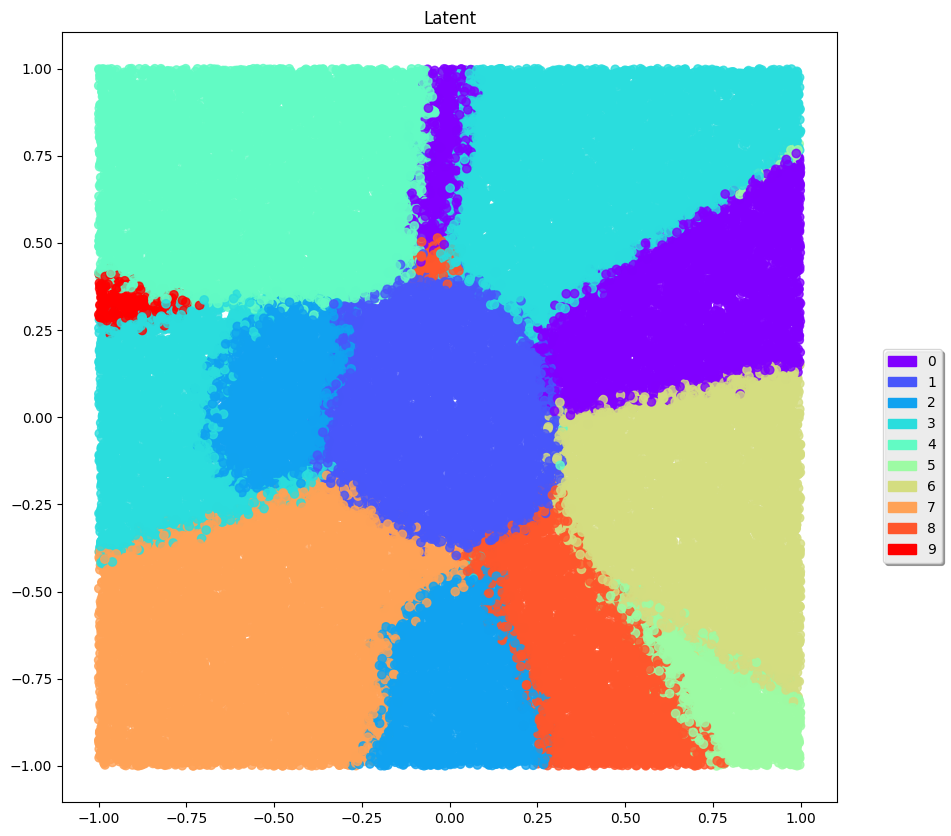}
\includegraphics[scale=0.11]{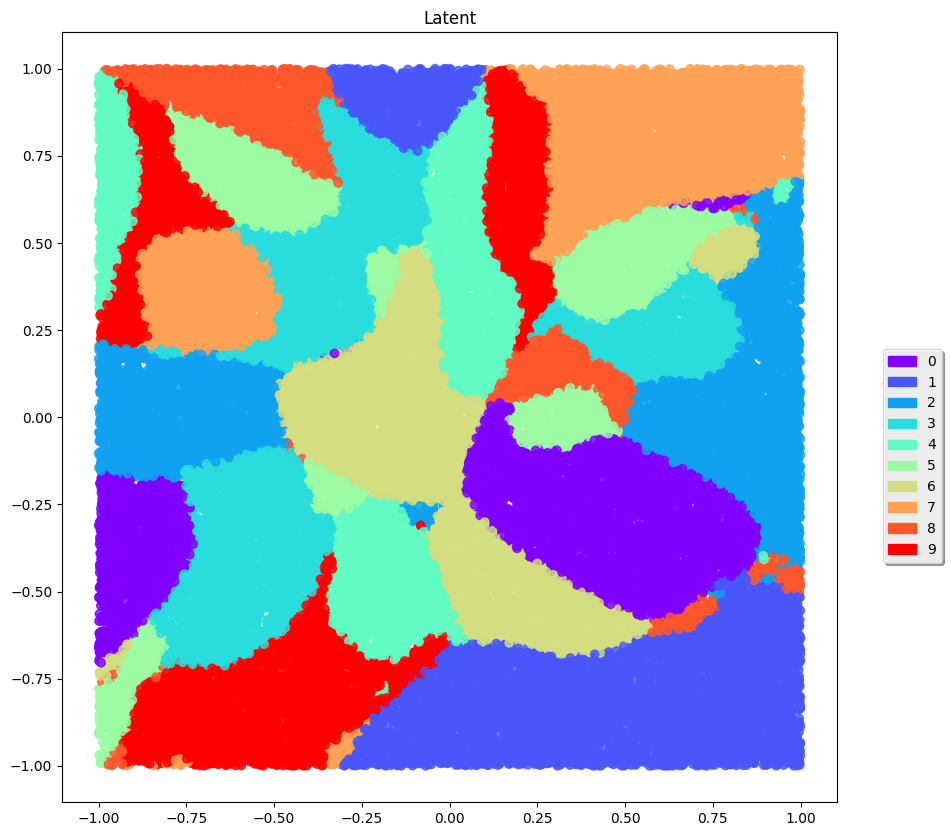}
\includegraphics[scale=0.11]{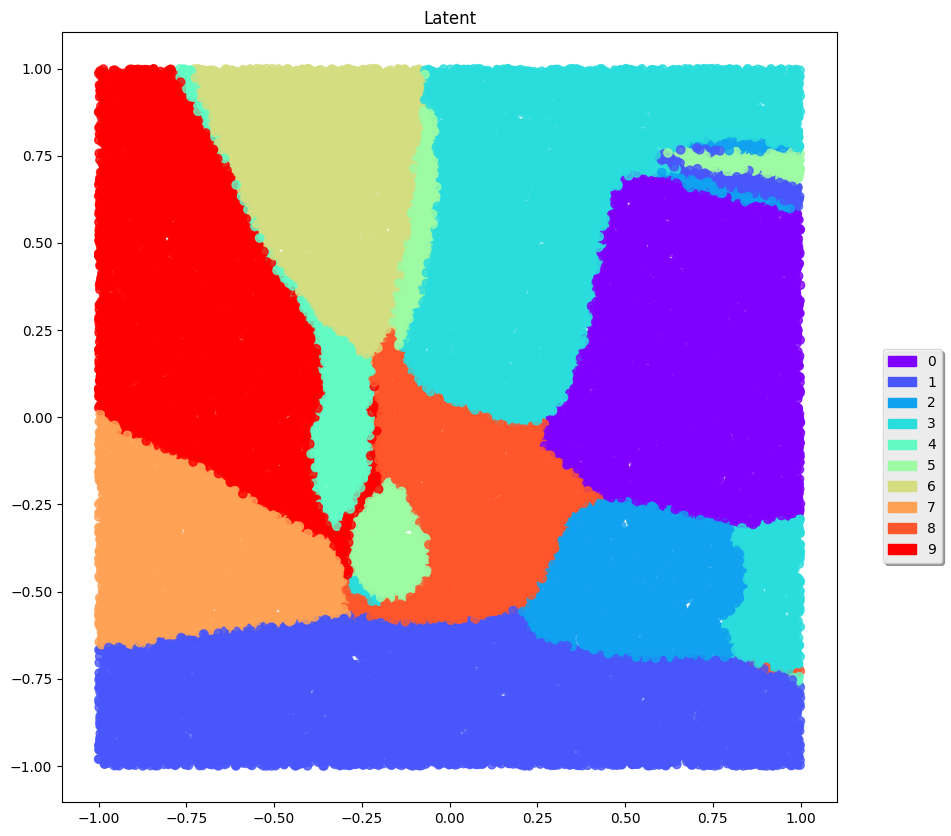}
\includegraphics[scale=0.11]{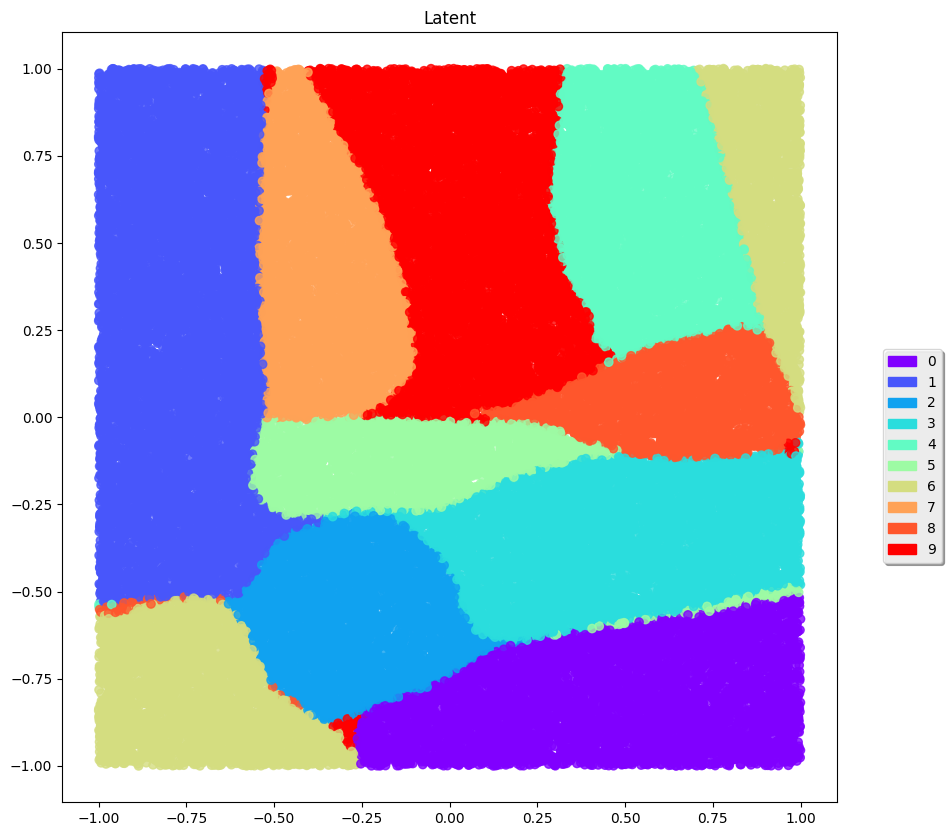}
\caption{Latent space visualization: The labels of 55K 2D latent variables obtained by (a) DCGAN, (b) WGANGP, (c) our Dist-GAN$_2$ (without latent-data distance) and (d) our Dist-GAN$_3$ (with our proposed latent-data distance). The Dist-GAN settings are defined in the section of Experimental Results.}
\label{classified_mnist_dcgan}
\end{figure*}
Mode collapse occurs when ``the generator collapses to a parameter setting where it always emits the same point. When collapse to a single mode is imminent, the gradient of the discriminator may point in similar directions for many similar points." \cite{salimans-nisp-2016}. 
Previous work usually examines mode collapse by visualizing a few collapsed samples (generated from random latent samples of a prior distribution). Fig. \ref{collapsed_mnist_dcgan}a is an example. However, the data space is high-dimensional, therefore it is difficult to visualize points in the data space. On the other hand, the latent space is lower-dimensional and controllable, and it is possible to visualize the entire 2D/3D spaces. 
Thus, it could be advantageous to examine mode collapse in the latent space.
However, the problem is that GAN is not invertible to map the data samples back to the latent space.
Therefore, we propose the following method to visualize the samples and examine mode collapse in the latent space.
We apply an off-the-shelf classifier. This classifier predicts labels of the generated samples. We visualize these class labels according to the latent samples, see
Fig. \ref{collapsed_mnist_dcgan}b.
This is possible because, for many datasets such as MNIST, pre-trained classifiers  can achieve high accuracy, \eg 0.04\% error rate. 
\subsection{Distance constraint: Motivation}
Fig. \ref{collapsed_mnist_dcgan}b is the latent sample visualization using  this technique, and the latent samples are uniformly distributed in a 2D latent space of $[-1, 1]$. 
Fig. \ref{collapsed_mnist_dcgan}b clearly suggests the extent of mode collapse: many latent samples  from large regions of latent space are collapsed into the same digit, \eg `1'.
Even some latent samples reside  very far apart from each other, they map to the same digit.
This suggests that  a generator $G_{\theta}$ with parameter $\theta$ has mode collapse when there are many latent samples mapped to small regions of the data space:
\begin{equation}
\mathrm{x_i} = G_{\theta}(\mathrm{z_i}), \mathrm{x_j} = G_{\theta}(\mathrm{z_j}): f(\mathrm{x_i}, \mathrm{x_j}) < \delta_{\mathrm{x}}
\label{mode_collapse_eqn_1}
\end{equation}
Here $\{\mathrm{z_i}\}$ are latent samples, and $\{\mathrm{x_i}\}$ are corresponding synthesized samples by $G_{\theta}$. $f$ is some distance metric in the data space, and $\delta_{\mathrm{x}}$ is a small threshold in the data space. Therefore, we propose to address mode collapse using a distance metric $g$ in latent space, and a small threshold $\delta_{\mathrm{z}}$ of this metric, to restrain $G_{\theta}$ as follows:
\begin{equation}
g(\mathrm{z_i}, \mathrm{z_j}) > \delta_{\mathrm{z}} \to f(\mathrm{x_i}, \mathrm{x_j}) > \delta_{\mathrm{x}}
\label{mode_collapse_eqn_2}
\end{equation}
However, determining good functions $f, g$ for two spaces of different dimensionality and their thresholds $\delta_{\mathrm{x}}, \delta_{\mathrm{z}}$ is not straightforward. Moreover, applying these constraints to GAN is not simple, because GAN has only one-way mapping from latent to data samples. In the next section, we will propose novel formulation to represent this constraint in latent-data distance and apply this to GAN.

We have also tried to apply this visualization for two state-of-the-art methods: DCGAN \cite{radford-arxiv-2015}, WGANGP \cite{gulrajani-arxiv-2017} on the MNIST dataset (using the code of \cite{gulrajani-arxiv-2017}). Note that all of our experiments were conducted in the unsupervised setting. The off-the-shelf classifier is used here to determine the labels of generated samples solely for visualization purpose. Fig. \ref{classified_mnist_dcgan}a and Fig. \ref{classified_mnist_dcgan}b represent the labels of the 55K latent variables of DCGAN and WGANGP respectively at iteration of 70K.  Fig. \ref{classified_mnist_dcgan}a reveals that DCGAN is partially collapsed,  as it generates very few digits `5' and `9' according to their latent variables near the bottom-right top-left corners of the prior distribution. In contrast, WGANGP does not have mode collapse, as shown in Fig. \ref{classified_mnist_dcgan}b.  However, for WGANGP, the latent variables corresponding to  each digit are fragmented in many sub-regions. It is an interesting observation for WGANGP. We will investigate this as our future work.
\subsection{Improving GAN using Distance Constraints}
We apply the idea of Eqn. \ref{mode_collapse_eqn_2} to improve generator through an AE. We apply AE to encode data samples into latent variables and use these encoded latent variables to direct the generator's mapping from the entire latent space. First, we train an  AE (encoder $E_{\omega}$ and decoder $G_\theta$), then we   train the discriminator $D_{\gamma}$ and the generator $G_{\theta}$. Here, the generator is the decoder of AE and $\omega, \theta, \gamma$ are the parameters of the encoder, generator, and discriminator respectively. Two main reasons for training an AE are: (i) to regularize the parameter $\theta$ at each training iteration, and (ii) to direct the generator to synthesize samples similar to real training samples. 
%Furthermore, knowing the arrangement of encoded data samples in the latent space allows us to better control the sampling of the generator. The idea will be formed as \textit{a data-latent distance constraint} for AE, and the objective is written:
We include an additional {\em latent-data distance constraint} to train the AE:
\begin{equation}
\min_{\omega,\theta} L_R(\omega,\theta) + \lambda_{\mathrm{r}} L_W(\omega,\theta)
\label{recon_01}
\end{equation}
where $L_R(\omega,\theta) = ||\mathrm{x} - G_{\theta}(E_{\omega}(\mathrm{x}))||_2^2$ is the conventional AE objective.
The latent-data distance constraint $L_W(\omega,\theta)$ is to regularize the generator and prevent it from being collapsed. This term will be discussed later. Here, $\lambda_{\mathrm{r}}$ is the constant. The reconstructed samples $G_{\theta}(E_{\omega}(\mathrm{x}))$ can be approximated by $G_{\theta}(E_{\omega}(\mathrm{x})) = \mathrm{x} + \varepsilon$, where $\varepsilon$ is the reconstruction error. Usually the capacity of $E$ and $G$ are large enough so that $\epsilon$ is small (like noise). Therefore, it is reasonable to consider those reconstructed samples as ``real" samples (plus noise $\varepsilon$). The pixel-wise reconstruction may cause blurry. To circumvent this, we instead use feature-wise distance \cite{larsen-arxiv-2015} or similarly feature matching \cite{salimans-nisp-2016}:  $L_R(\omega,\theta) = ||\Phi(\mathrm{x}) - \Phi(G_{\theta}(E_{\omega}(\mathrm{x})))||_2^2$. Here
$\Phi(\mathrm{x})$ is the high-level feature obtained from some middle layers of deep networks. In our implementation, $\Phi(\mathrm{x})$ is the feature output from the last convolution layer of discriminator $D_{\gamma}$. Note that in the first iteration, the parameters of discriminator are randomly initialized, and features produced from this discriminator is used to train the AE. 

\begin{figure}[t]
\centering
\includegraphics[scale=0.35]{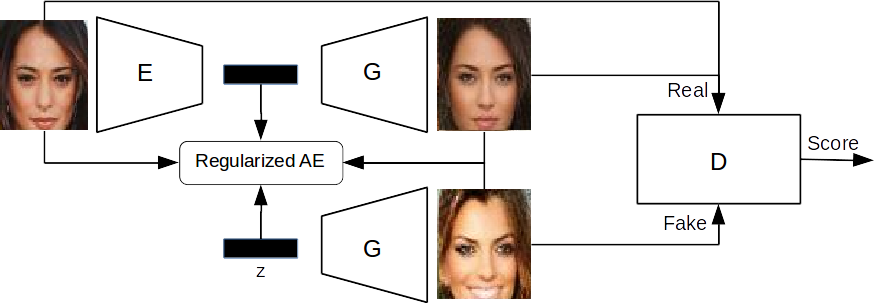}
\caption{The architecture of Dist-GAN includes Encoder (E), Generator (G) and Discriminator (D). Reconstructed samples are considered as ``real''. The input, reconstructed, and generated samples as well as the input noise and encoded latent are all used to form the latent-data distance constraint for AE (regularized AE).}
\label{architecture-diagram}
\end{figure}

Our framework is shown in Fig. \ref{architecture-diagram}.  We propose to train encoder $E_{\omega}$, generator $G_{\theta}$ and discriminator $D_{\gamma}$ following the order: (i) fix $D_{\gamma}$ and train $E_{\omega}$ and $G_{\theta}$ to minimize the reconstruction loss Eqn. \ref{recon_01} (ii) fix $E_{\omega}$, $G_{\theta}$, and train $D_{\gamma}$ to minimize (Eqn. \ref{gan_objective_02}), and (iii) fix $E_{\omega}$, $D_{\gamma}$ and train $G_{\theta}$ to minimize (Eqn. \ref{gan_objective_01}).
\subsubsection{Generator and discriminator objectives}
When training the generator, maximizing the conventional generator objective $\mathbb{E}_{\mathrm{z}} \sigma (D_\gamma(G_\theta(\mathrm{z})))$ \cite{goodfellow-nisp-2014} tends to produce samples at high-density modes, and this leads to mode collapse easily. Here, $\sigma$ denotes the sigmoid function and $\mathbb{E}$ denotes the expectation. Instead, we train the generator with our proposed ``discriminator-score distance". We align the  synthesized sample distribution to real sample distribution with the $\ell_1$ distance.  The alignment is through the discriminator score, see Eqn. \ref{gan_objective_01}. Ideally, the generator synthesizes samples similar to the samples drawn from the real distribution, and this also helps reduce missing mode issue.
\begin{equation}
%\normalsize
% Store the current equation number.
\min_{\theta}\mathcal{L}_G(\theta) = |\mathbb{E}_{\mathrm{x}} \sigma (D_\gamma(\mathrm{x})) - \mathbb{E}_{\mathrm{z}} \sigma (D_\gamma(G_\theta(\mathrm{z})))|
\label{gan_objective_01}
\end{equation}
The objective function of the discriminator is shown in Eqn. \ref{gan_objective_02}. It is different from original discriminator of GAN in two aspects. First, we indicate the reconstructed samples as ``real", represented by the term $L_C = \mathbb{E}_{\mathrm{x}} \log \sigma (D_\gamma(G_\theta(E_\omega(\mathrm{x}))))$. Considering the reconstructed samples as ``real" can  systematically slow down the convergence of discriminator, so that the gradient from discriminator is not saturated too quickly. {\em In particular, the convergence of the discriminator is coupled with the convergence of AE}. This is  an important constraint. In contrast, if we consider the reconstruction as ``fake" in our model, this speeds up the discriminator convergence, and the discriminator converges faster than both generator and encoder. This leads to gradient saturation of $D_\gamma$. 
%Fig. \ref{mode_collapse_eqn_1} is some result that shows serious mode collapse in this case with the same input of 2D latent as above. 
Second, we apply the gradient penalty $L_P = (||\nabla_{\hat{\mathrm{x}}} D_\gamma(\hat{\mathrm{x}})||_2^2 - 1)^2$ for the discriminator objective (Eqn. \ref{gan_objective_02}), where $\lambda_{\mathrm{p}}$ is penalty coefficient, and $\hat{\mathrm{x}} = \epsilon \mathrm{x} + (1 - \epsilon) G(\mathrm{z})$, $\epsilon$ is a uniform random number $\epsilon \in U[0,1]$. This penalty was used to enforce Lipschitz constraint of Wasserstein-1 distance \cite{gulrajani-arxiv-2017}. In this work, we also find this useful for JS divergence and stabilizing our model. It should be noted that  using this gradient penalty alone 
cannot solve the convergence issue, similar to WGANGP. The problem is partially solved when combining this with our proposed generator objective in Eqn. \ref{gan_objective_01}, i.e., discriminator-score distance. However, the problem cannot be completely solved,  \eg mode collapse on MNIST dataset with 2D latent inputs as shown in Fig. \ref{classified_mnist_dcgan}c. Therefore, we apply the proposed latent-data distance constraints as additional regularization term for AE:  $L_W(\omega,\theta)$, to be discussed in the next section.
\begin{equation}
\begin{split}
\min_{\gamma}\mathcal{L}_D(\omega,\theta,\gamma) & = -(\mathbb{E}_{\mathrm{x}} \log \sigma (D_\gamma(\mathrm{x})) + \mathbb{E}_{\mathrm{z}} \log(1 - \sigma (D_\gamma(G_\theta(\mathrm{z})))) \\
&+ \mathbb{E}_{\mathrm{x}} \log \sigma (D_\gamma(G_\theta(E_\omega(\mathrm{x})))) - \lambda_{\mathrm{p}}\mathbb{E}_{{\hat{\mathrm{x}}}}(||\nabla_{\hat{\mathrm{x}}} D_\gamma(\hat{\mathrm{x}})||_2^2 - 1)^2)
\end{split}
\label{gan_objective_02}
\end{equation}

\subsubsection{Regularizing Autoencoders by Latent-Data Distance Constraint}
\begin{figure*}
\centering
\includegraphics[scale=0.21]{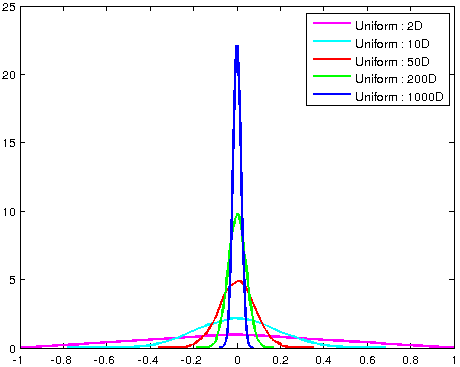}
\includegraphics[scale=0.21]{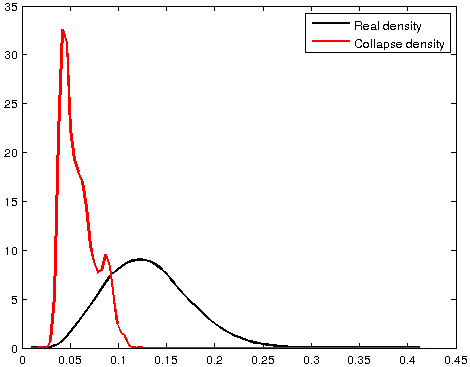}
\includegraphics[scale=0.21]{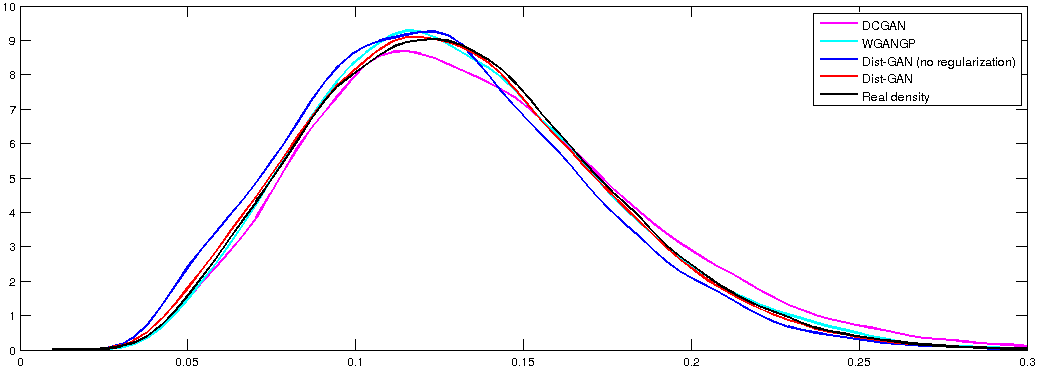}
\caption{(a) The 1D frequency density of outputs using $\mathrm{M_d}$ from 
uniform distribution of different dimensionality. (b) One example of the density when mode collapse occurs. (c) The 1D density of real data and generated data obtained by different methods: DCGAN (kldiv: 0.00979), WGANGP (kldiv: 0.00412), Dist-GAN$_2$ (without data-latent distance constraint of AE, kldiv: 0.01027), and Dist-GAN (kldiv: 0.00073).}
\label{1d-density}
\end{figure*}

In this section, we discuss the latent-data distance constraint $L_W(\omega,\theta)$ to regularize AE in order to reduce mode collapse in the generator (the decoder in the AE).
%First, we observe the encoded latents obtained by our Dist-GAN$_2$ model ($\lambda_{\mathrm{r}} = 0$ or without data-latent distance constraint) on MNIST dataset via the same technique as in Fig. \ref{classified_mnist_dcgan}a, \ref{classified_mnist_dcgan}b. Its 55K latent codes are shown in Fig. \ref{classified_mnist_dcgan}c. Although Dist-GAN$_2$ worked pretty well on synthetic data (Section \ref{synthetic_data_experiment}), this figure reveals it is still partially collapsed, \eg small regions of latent variables generating digits `4' and `5'. The areas covered by MNIST digits are significantly different in this space. If we use smaller network architectures, the mode collapse is even more serious. Intuitively, in order to avoid this problem, AE needs to assign more spaces for some digits, that allows random noise $\mathrm{z}$ falling into those digits' regions with higher probability.
In particular, we use noise input to constrain encoder's outputs, and simultaneously reconstructed samples to constrain the generator's outputs. Mode collapse occurs when the generator synthesizes  low diversity of samples in the data space given different latent inputs. Therefore, to reduce mode collapse, we aim to achieve:  if the distance of any two latent variables $g(\mathrm{z_i}, \mathrm{z_j})$ is small (large) in the latent space, the corresponding distance $f(\mathrm{x_i}, \mathrm{x_j})$ in data space should be small (large), and vice versa. We propose a latent-data distance regularization $L_W(\omega,\theta)$: 
\begin{equation}
L_W(\omega,\theta) = ||f(\mathrm{x},G_\theta(\mathrm{z})) - \lambda_{\mathrm{w}}g(E_\omega(\mathrm{x}),\mathrm{z})||_2^2
\label{ae_regularization}
\end{equation}
where $f$ and $g$ are distance functions computed in data and latent space. $\lambda_{\mathrm{w}}$ is the scale factor due to the difference in dimensionality. 
It is not straight forward to compare distances in spaces of different dimensionality. Therefore, instead of using the direct distance functions, \eg Euclidean, $\ell_1$-norm, etc, we propose to  compare the matching score   $f(\mathrm{x},G_\theta(\mathrm{z}))$ of real and fake distributions,  and the matching score $g(E_\omega(\mathrm{x}),\mathrm{z})$ of two latent distributions. We use means as the matching scores. Specifically:
\begin{equation}
f(\mathrm{x},G_\theta(\mathrm{z})) = \mathrm{M_d}(\mathbb{E}_{\mathrm{x}}G_\theta(E_\omega(\mathrm{x})) - \mathbb{E}_{\mathrm{z}} G_\theta(\mathrm{z}))
\end{equation}
\begin{equation}
g(E_\omega(\mathrm{x}),\mathrm{z}) = \mathrm{M_d}(\mathbb{E}_{\mathrm{x}}E_\omega(\mathrm{x}) - \mathbb{E}_{\mathrm{z}}\mathrm{z})
\end{equation}
where $\mathrm{M_d}$ computes the average of all dimensions of the input. Fig. \ref{1d-density}a illustrates 1D frequency density of 10000 random samples mapped by $\mathrm{M_d}$ from $[-1,1]$ uniform distribution of different dimensionality.
We can see that outputs of $\mathrm{M_d}$ from high dimensional spaces 
have small values. 
Thus, we require $\lambda_\mathrm{w}$ in 
(\ref{ae_regularization}) to account for the difference in dimensionality.
%comparing  two different densities of data and latent samples requires $\lambda_\mathrm{w}$. 
Empirically, we found $\lambda_\mathrm{w} = \sqrt{\frac{d_\mathrm{z}}{d_\mathrm{x}}}$ suitable, where $d_\mathrm{z}$ and $d_\mathrm{x}$ are dimensions of latent and data samples respectively. 
%Intuitively, $\mathcal{W}(\omega,\theta)$ can enforce the relative distance between two manifolds. When the generator is collapsed, $G_\theta(\mathrm{z})$ always produces similar outputs for many given inputs $\mathrm{z}$ that leads $f(\mathrm{x},G_\theta(\mathrm{z}))$ to be small. Otherwise, $\mathrm{z}$ is random and $E_\omega(\mathrm{x})$ are encoded from data. Hence, it's difficult for $g(E_\omega(\mathrm{x}),\mathrm{z})$ to match the score of $f(\mathrm{x},G_\theta(\mathrm{z}))$ to minimize the data-latent distance constraint objective. As a result, enforcing $f$ and $g$ can reduce drastically the mode collapse problem. 
Fig. \ref{1d-density}b shows the frequency density of a collapse mode case. We can observe that  the 1D density of generated samples is clearly different from that of the real data. Fig. \ref{1d-density}c compares 1D frequency densities of 55K MNIST samples generated by different methods. Our Dist-GAN method can estimate better 1D density than DCGAN and WGANGP measured by KL divergence (kldiv) between the densities of generated samples and real samples. 
%This 1D mapping is not bijective. Hence, it is likely not always true that our method can approximate better data distribution than others. However, we see empirically that the generated 1D density is far from the real data density, it is more likely in collapse mode. It is a good evidence along with later empirical experiments to believe that our method is better than others in term of approximating the data distribution and handling the mode collapse issue. 

The entire algorithm is presented in Algorithm. \ref{alg-01}.
\begin{algorithm*}
 \footnotesize
 \caption{Dist-GAN}
 \begin{algorithmic}[1]
 \STATE Initialize discriminators, encoder and generator $D_\gamma, E_\omega, G_\theta$
 \REPEAT
  \STATE $\mathrm{x^m} \leftarrow$ Random minibatch of $m$ data points from dataset.
  \STATE $\mathrm{z^m} \leftarrow$ Random $m$ samples from noise distribution $P_\mathrm{z}$
  \STATE // \textit{Training encoder and generator using $\mathrm{x^m}$ and $\mathrm{z^m}$ by Eqn. \ref{recon_01}}
  \STATE $\omega, \theta \leftarrow \min_{\omega, \theta} 
  L_R(\omega, \theta) + \lambda_{\mathrm{r}} L_W(\omega, \theta)$
  %\STATE // \textit{Interpolate latent variables by Eqn. \ref{interpolate_latent}}
  %\STATE $\mathrm{z^m_i} \leftarrow (1 - \alpha)\mathrm{E_\omega(\mathrm{x^m})} + \alpha \mathrm{z^m}$ // \textit{where $\alpha \in [0,1]$}
  \STATE // \textit{Training discriminators according to Eqn. \ref{gan_objective_02} on $\mathrm{x^m}, \mathrm{z^m}$} %, \mathrm{z^m_i}$}
  \STATE $\gamma \leftarrow \min_{\gamma} \mathcal{L}_D(\omega, \theta, \gamma)$  
  \STATE // \textit{Training the generator on $\mathrm{x^m}, \mathrm{z^m}$  according to Eqn. \ref{gan_objective_01}.} % by Eqn. \ref{objective_G_02}}
  \STATE $\theta \leftarrow \min_{\theta} \mathcal{L}_G(\theta)$  %+ \mathbb{E}_{\mathrm{z_i}}\log(D_\gamma(G_\theta(\mathrm{z^m_i})))$ 
  %\STATE $\alpha = \alpha * \nu^{\mathrm{epoch}}$ 
 \UNTIL
 \RETURN $E_\omega, G_\theta, D_\gamma$ 
 \end{algorithmic} 
 \label{alg-01}
\end{algorithm*}

\section{Experimental Results}
%In this section, we show that our method can improve the mode coverage on our synthetic datasets and the MNIST datasets, and generate quality faces on CelebA dataset. We also compute standard scores and compare to the state of the art on MNIST-1K, CelebA, CIFAR-10 and STL-10. All experiments are conducted in the unsupervised setting.
\subsection{Synthetic data}
\label{synthetic_data_experiment}
All our experiments are conducted using the unsupervised setting.
First, we use  synthetic data to evaluate how well our Dist-GAN can approximate the data distribution. We use a synthetic dataset of 25 Gaussian modes in grid layout similar to \cite{dumoulin-arxiv-2016}. Our dataset contains 50K training points in 2D, and we draw 2K generated samples for testing. For fair comparisons, we use equivalent architectures and setup for all methods in the same experimental condition if possible. The architecture and network size are similar to \cite{metz-arxiv-2016} on the 8-Gaussian dataset, except that we use one more hidden layer. We use fully-connected layers and Rectifier Linear Unit (ReLU) activation for input and hidden layers, sigmoid for output layers. The network size of encoder, generator and discriminator are presented in Table 1 of Supplementary Material, where $d_\mathrm{in} = 2$, $d_\mathrm{out} = 2$, $d_\mathrm{h} = 128$ are dimensions of input, output and hidden layers respectively. $N_\mathrm{h} = 3$ is the number of hidden layers. The output dimension of the encoder is the dimension of the latent variable. Our prior distribution is  uniform $[-1,1]$. We use Adam optimizer with learning rate $\mathrm{lr} = 0.001$, and the exponent decay rate of first moment $\beta_1 = 0.8$. The learning rate is decayed every $10K$ steps with a base of $0.9$. The mini-batch size is $128$. The training stops after 500 epochs. To have fair comparison, we carefully fine-tune other methods (and use weight decay during training if this achieves better results) to ensure they achieve their best results on the synthetic data. For evaluation, a mode is missed if there are less than 20 generated samples registered into this mode, which is measured by its mean and variance of $0.01$ \cite{li-arxiv-2017,metz-arxiv-2016}. A method has mode collapse if there are missing modes. In this experiment, we fix the parameters $\lambda_{\mathrm{r}} = 0.1$ (Eqn. \ref{recon_01}), $\lambda_{\mathrm{p}} = 0.1$ (Eqn. \ref{gan_objective_02}), $\lambda_{\mathrm{w}} = 1.0$ (Eqn. \ref{ae_regularization}). For each method, we repeat eight runs and report the average.
\begin{figure*}
\centering
\includegraphics[scale=0.083]{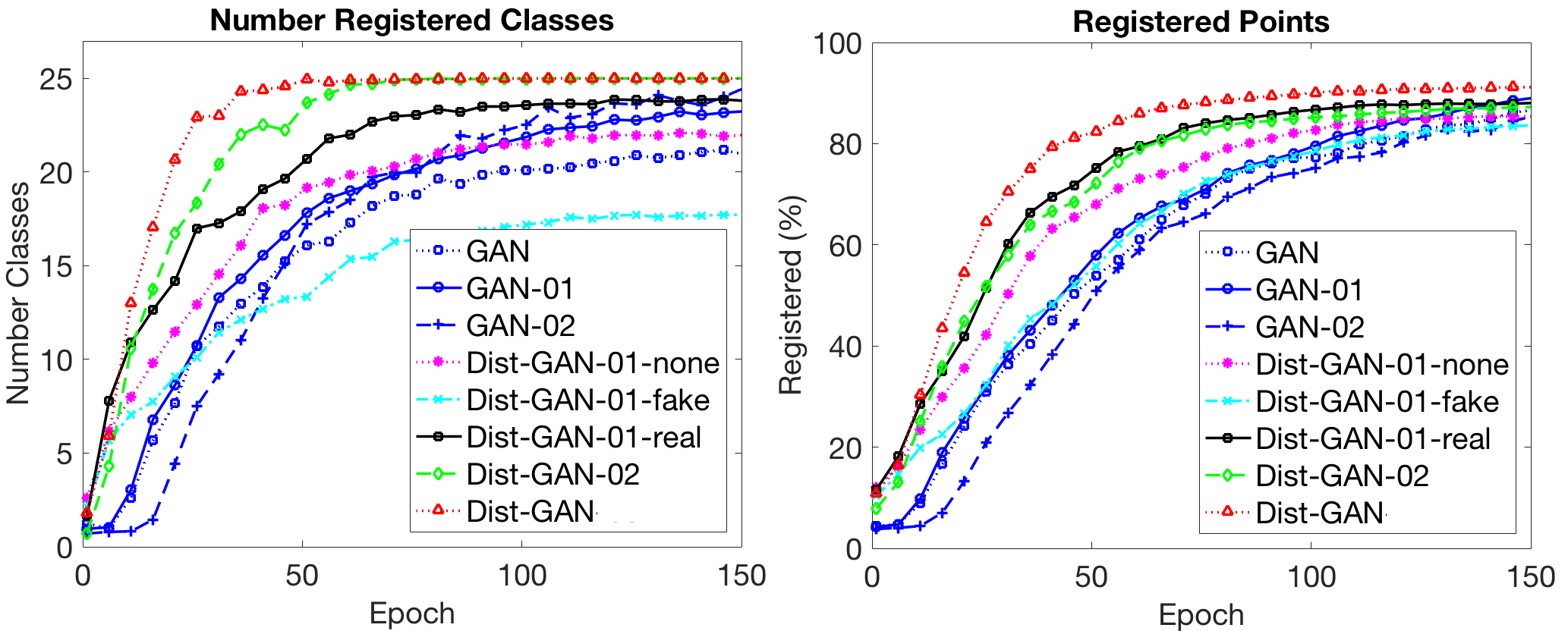}
\includegraphics[scale=0.15]{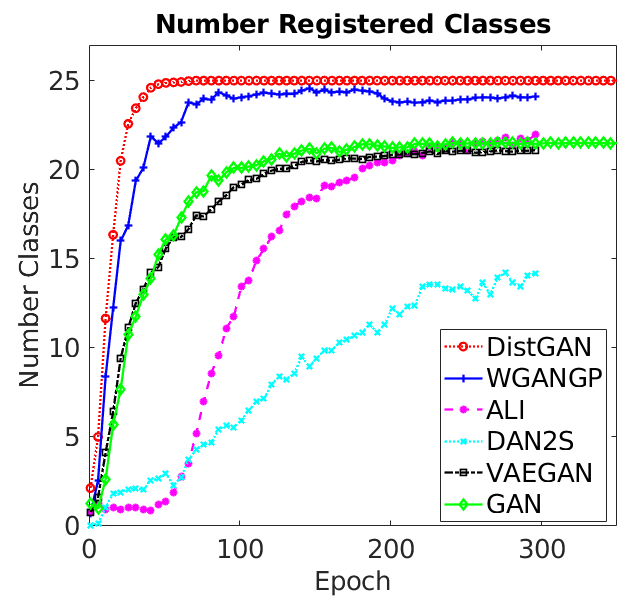}
\includegraphics[scale=0.15]{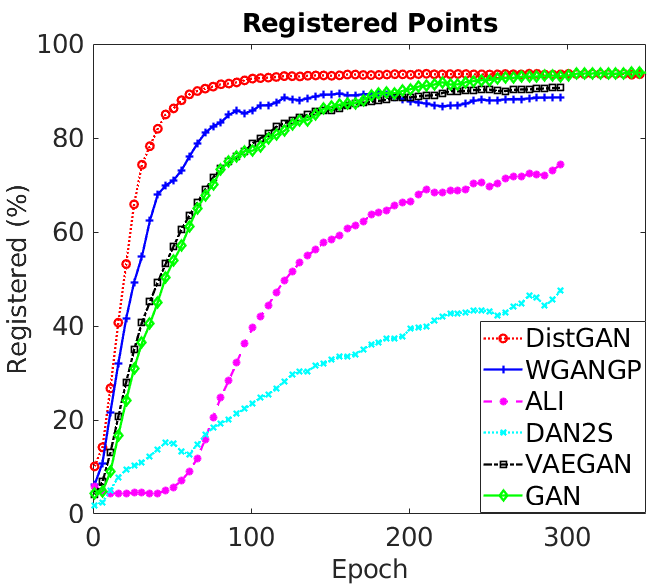}
\caption{From left to right figures: (a), (b), (c), (d). The number of registered modes (a) and points (b) of our method with two different settings on the synthetic dataset. We compare our Dist-GAN to the baseline GAN \cite{goodfellow-nisp-2014} and other methods on the same dataset measured by the number of registered modes (classes) (c) and points (d).}
\label{toydata_comparison_01}
\end{figure*}

First, we highlight the capability of our model to approximate the distribution $P_\mathrm{x}$ of synthetic data. We carry out the ablation experiment to understand the influence of each proposed component with different settings:
\begin{itemize}
\item Dist-GAN$_1$: uses the ``discriminator-score distance" for generator objective ($\mathcal{L}_G$) and the AE loss $L_R$ but does not use data-latent distance constraint term ($L_W$) and gradient penalty ($L_P$). This setting has three different versions as using reconstructed samples ($L_C$) as ``real", ``fake" or ``none" (not use it) in the discriminator objective. 
\item Dist-GAN$_2$: improves from Dist-GAN$_1$ (regarding reconstructed samples as ``real") by adding the gradient penalty $L_P$.
\item Dist-GAN: improves the Dist-GAN$_2$ by adding the data-latent distance constraint $L_W$. (See Table 3 in Supplementary Material for details).
\end{itemize}

The quantitative results are shown in Fig. \ref{toydata_comparison_01}. Fig. \ref{toydata_comparison_01}a is the number of registered modes changing over the training. Dist-GAN$_1$ misses a few modes while Dist-GAN$_2$ and Dist-GAN generates all 25 modes after about 50 epochs. Since they almost do not miss any modes, it is reasonable to compare the number of registered points as in Fig. \ref{toydata_comparison_01}b. Regarding reconstructed samples as ``real" achieves better results than regarding them as  ``fake" or ``none". 
%If ``fake" reconstructed samples are used, the model gets stuck since the gradient of discriminator is saturated and misses many modes. That highlights our choice of using ``real" reconstructed samples. 
It is reasonable that Dist-GAN$_1$ obtains similar results as the baseline GAN when not using the reconstructed samples in discriminator objective (``none" option). Other results show the improvement when adding the gradient penalty into the discriminator (Dist-GAN$_2$). Dist-GAN demonstrates the effectiveness of using the proposed latent-data constraints, when comparing with Dist-GAN$_2$. 

To highlight the effectiveness of our proposed ``discriminator-score distance" for the generator, we use it to improve the baseline GAN \cite{goodfellow-nisp-2014}, denoted by GAN$_1$. Then, we propose GAN$_2$ to improve GAN$_1$ by adding the gradient penalty. We can observe that  combination of our proposed generator objective and gradient penalty can improve stability of  GAN.
% stable, there is a slight mode collapse. In contrast, our best Dist-GAN settings can recover all modes and converge faster than accordingly improved GAN versions.
\begin{figure}[t]
\centering
\includegraphics[scale=0.40]{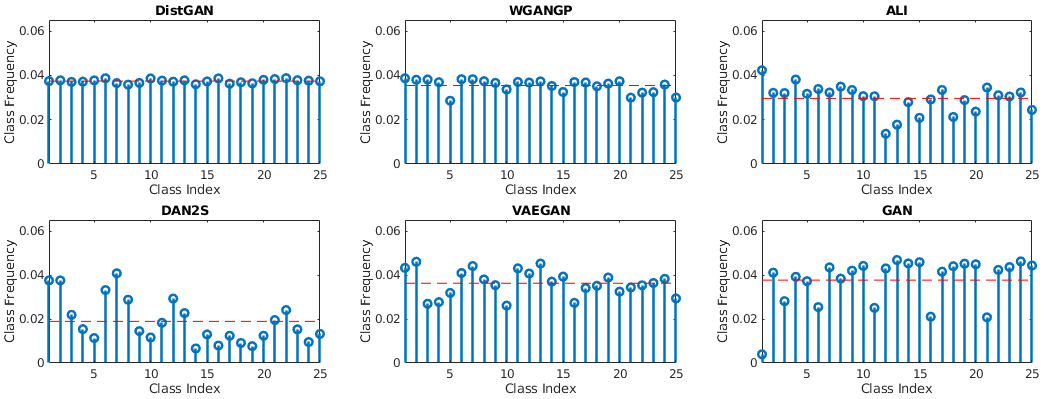}
\caption{The mode balance obtained by different methods.}
\label{toydata_frequency}
\end{figure}
We compare our best setting (Dist-GAN) to previous work. ALI \cite{dumoulin-arxiv-2016} and DAN-2S \cite{li-arxiv-2017} are recent works using encoder/decoder in their model. VAE-GAN \cite{larsen-arxiv-2015} introduces a similar model. WGAN-GP \cite{gulrajani-arxiv-2017} is one of the current state of the art. The numbers of covered modes and registered points are presented in Fig. \ref{toydata_comparison_01}c and Fig \ref{toydata_comparison_01}d respectively. The quantitative numbers of last epochs are shown in Table 2 of Supplementary Material. In this table, we report also Total Variation scores to measure the mode balance. The result for each method is the average of eight runs. Our method outperforms GAN \cite{goodfellow-nisp-2014}, DAN-2S \cite{li-arxiv-2017}, ALI \cite{dumoulin-arxiv-2016}, and VAE/GAN \cite{larsen-arxiv-2015} on the number of covered modes. While WGAN-GP sometimes misses one mode and diverges, our method (Dist-GAN) does not suffer from  mode collapse in all eight runs. Furthermore, we achieve a higher number of registered samples than WGAN-GP and all others. Our method is also better than the rest with Total Variation (TV) \cite{li-arxiv-2017}. Fig. \ref{toydata_frequency} depicts the detail proportion of generated samples of 25 modes. (More visualization of generated samples in Section 2 of Supplementary Material).
\subsection{MNIST-1K}
For image datasets, we use $\Phi(\mathrm{x})$ instead $\mathrm{x}$ for the reconstruction loss and the latent-data distance constraint in order to avoid the blur. We fix the parameters $\lambda_{\mathrm{p}} = 1.0$, and $\lambda_{\mathrm{r}} = 1.0$ for all image datasets that work consistently well. The $\lambda_{\mathrm{w}}$ is automatically computed from dimensions of features $\Phi(\mathrm{x})$ and latent samples.
\begin{figure*}[t]
\centering
\includegraphics[scale=0.42]{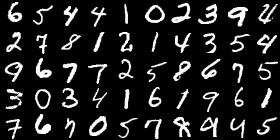}
\includegraphics[scale=0.42]{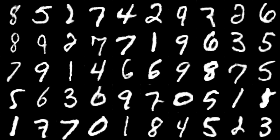}
\includegraphics[scale=0.19]{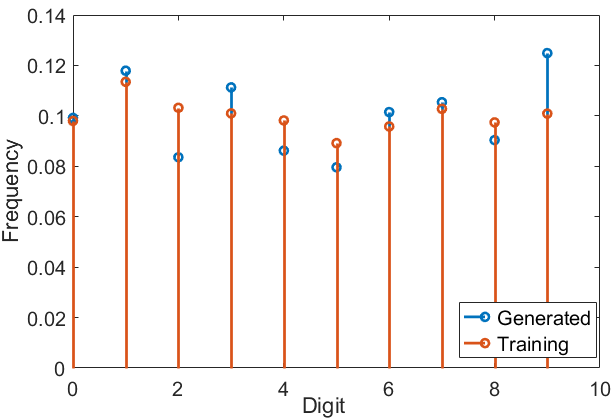}
\caption{The real and our generated samples in one mini-batch. And the number of generated samples per class obtained by our method on the MNIST dataset. We compare our frequency of generated samples to the ground-truth via KL divergence: KL = 0.01.}
\label{mnist_comparison_01}
\end{figure*}
Our model implementation for MNIST uses the published code of WGAN-GP \cite{gulrajani-arxiv-2017}. Fig. \ref{mnist_comparison_01} from left to right are the real samples, the generated samples and the  frequency of each digit generated by our method for standard MNIST. It demonstrates that our method can approximate well the MNIST digit distribution. Moreover, our generated samples look realistic with different styles and strokes that resemble the real ones. In addition, we follow the procedure in  \cite{metz-arxiv-2016} to construct a more challenging 1000-class MNIST (MNIST-1K) dataset. 
%This dataset is built by stacking three random digits to form an RGB image (one digit per a channel). 
It has 1000 modes from 000 to 999. 
%The digits are classified by a pre-train classifier (0.4\% in our case). 
We create a total of 25,600 images. We compare methods by counting the number of covered modes (having at least one sample \cite{metz-arxiv-2016}) and computing KL divergence. To be fair, we adopt the equivalent network architecture (low-capacity generator and two crippled discriminators K/4 and K/2) as proposed by \cite{metz-arxiv-2016}. Table \ref{mnist1k_comparison_01} presents the number of modes and KL divergence of compared methods. Results show that our method outperforms all others in the number of covered modes, especially with the low-capacity discriminator (K/4 architecture), where our method has  150 modes more than the second best. Our method reduces the gap between the two architectures (\eg about 60 modes), which is smaller than other methods. For both architectures, we obtain better results for both KL divergence and the number of recovered modes. All results support that our proposed Dist-GAN handles better mode collapse, and is robust even in case of imbalance in generator and discriminator.
\begin{table*}
\centering
\scriptsize
\caption{The comparison on MNIST-1K of methods. We follow the setup and network architectures from Unrolled GAN.}
\begin{tabular}{c | c | c | c | c}
\textbf{Architecture} & \textbf{GAN} & \textbf{Unrolled GAN} & \textbf{WGAN-GP} & \textbf{Dist-GAN}\\ 
\hline
K/4, \# & 30.6 $\pm$ 20.7 & 372.2 $\pm$ 20.7 & 640.1 $\pm$ 136.3 & 859.5 $\pm$ 68.7 \\ 
K/4, KL & 5.99 $\pm$ 0.04 & 4.66 $\pm$ 0.46 & 1.97 $\pm$ 0.70 & 1.04 $\pm$ 0.29\\
\hline
K/2, \# & 628.0 $\pm$ 140.9 & 817.4 $\pm$ 39.9 & 772.4 $\pm$ 146.5 & 917.9 $\pm$ 69.6 \\ 
K/2, KL & 2.58 $\pm$ 0.75 & 1.43 $\pm$ 0.12 & 1.35 $\pm$ 0.55  & 1.06 $\pm$ 0.23
\end{tabular}
\label{mnist1k_comparison_01}
\end{table*}
\section{CelebA, CIFAR-10 and STL-10 datasets}
\begin{figure*}[t]
\centering
\includegraphics[scale=0.14]{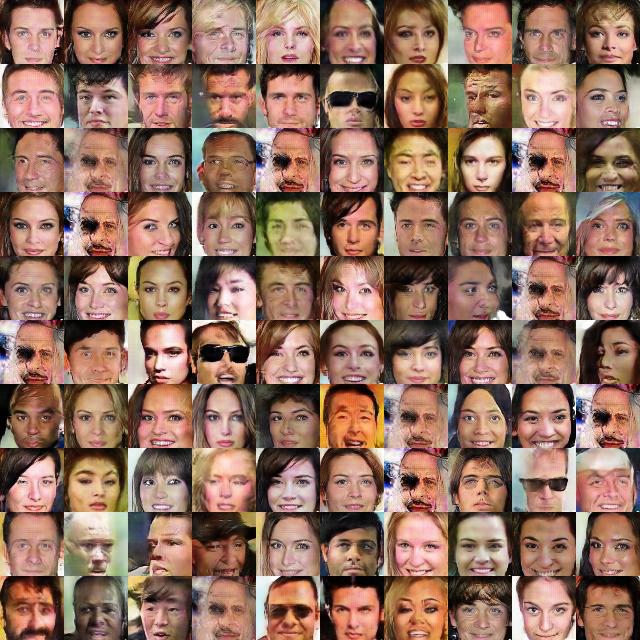}
\includegraphics[scale=0.14]{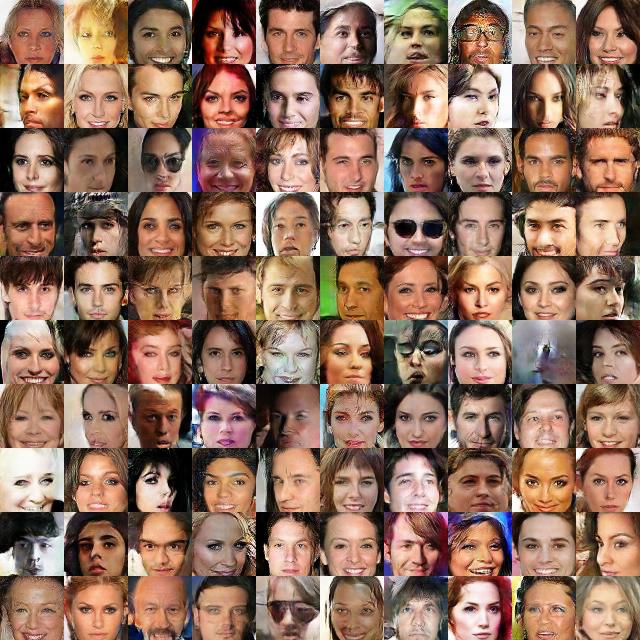}
\includegraphics[scale=0.14]{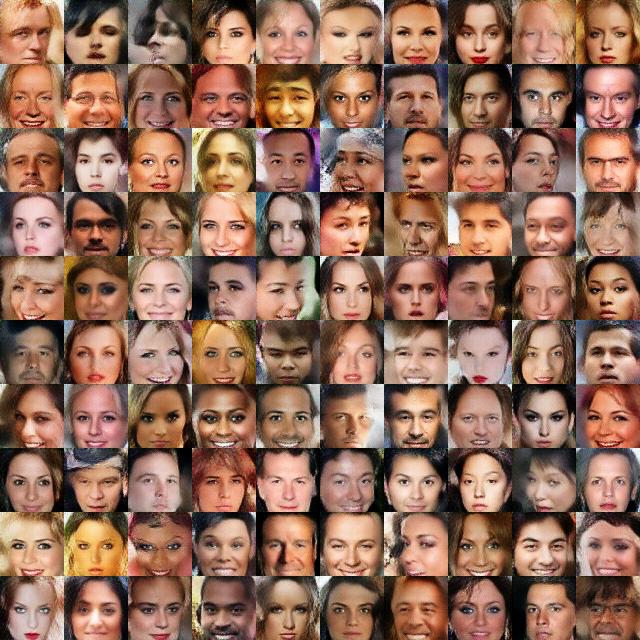}
\caption{Generated samples of DCGAN (50 epochs, results from \cite{gan_source}), WGAN-GP (50 epochs, results from \cite{gan_source}) and our Dist-GAN (50 epochs).}
\label{celeba_demo}
\end{figure*}
\begin{table*}[t]
\centering
\scriptsize
\caption{Comparing FID score to other methods. First two rows (CelebA, CIFAR-10) follow the experimental setup of \cite{lucic-2017-arxiv}, and the second row follow the experimental setup of \cite{miyato-iclr-2018} with standard CNN architectures, and the last row is with ResNet architecture.}
\begin{tabular}{c | c | c | c | c | c | c | c}
           & \textbf{NS GAN} & \textbf{LSGAN} & \textbf{WGANGP} & \textbf{BEGAN} & \textbf{VAEGAN} & \textbf{SN-GAN} & \textbf{Dist-GAN} \\
\hline
\textbf{CelebA}    & $58.0 \pm 2.7$  & $53.6 \pm 4.2$ & $26.8 \pm 1.2$  & $38.1 \pm 1.1$ & $27.5
 \pm 1.9$ & - & $23.7\pm 0.3$  \\
\textbf{CIFAR-10}     & $58.6 \pm 2.1$ & $67.1 \pm 2.9$ & $52.9 \pm 1.3$ &  $71.4 \pm 1.1$ & $58.1	\pm 3.2$ & - & $45.6 \pm 1.2$  \\
\hline
\textbf{CIFAR-10}     & - & - & - &  - & - & 29.3 & 28.23 \\
\textbf{CIFAR-10} (hinge) & - & - & - & - & - & 25.5 & 22.95 \\
\textbf{STL-10} (hinge) & - & - & - & - & - & 43.2 & 36.19\\
\hline
\textbf{CIFAR-10} (ResNet) & - & - & - & - & - & 21.70 $\pm$ .21 & 17.61 $\pm$ .30 \\
\end{tabular}
\label{fid_score}
\end{table*}
Furthermore, we use CelebA dataset and compare with DCGAN \cite{radford-arxiv-2015} and WGAN-GP \cite{gulrajani-arxiv-2017}. Our implementation is based on the open source \cite{gan_source_b,gan_source}.
%, which provided some available qualitative results of DCGAN and WGAN-GP. We choose this source because we want to show the robustness of our model: DCGAN gets collapsed but our model is robust. 
Fig. \ref{celeba_demo} shows samples generated by DCGAN, WGANGP and our Dist-GAN. While DCGAN is slightly collapsed at epoch 50, and WGAN-GP sometimes generates broken faces.
% (look unlike human faces). 
Our method does not suffer from such issues and can generate recognizable and realistic faces.
We also report  results for the CIFAR-10 dataset using DCGAN architecture \cite{radford-arxiv-2015} of same published code \cite{gulrajani-arxiv-2017}. The generated samples with our method trained on this dataset can be found in Section 4 of Supplementary Material. For quantitative results, we report the FID scores \cite{heusel-arxiv-2017} for both datasets. FID can detect intra-class mode dropping, and measure the diversity and the quality of generated samples. We follow the experimental procedure and model architecture in \cite{lucic-2017-arxiv}.
%, except the difference from \cite{lucic-2017-arxiv} is that we only use the default parameter settings of our method rather than performing the wide range hyper-parameter search for the best setting. However, 
Our method outperforms others for both CelebA and CIFAR-10, as shown in the first and second rows of Table \ref{fid_score}. Here, the results of other GAN methods are from \cite{lucic-2017-arxiv}. We also report FID score of VAEGAN on these datasets. Our method is better than VAEGAN. Note that we have also tried MDGAN, but it has serious mode collapsed for both these datasets. Therefore, we do not report its result in our paper. 

Lastly, we compare our model with recent SN-GAN \cite{miyato-iclr-2018} on CIFAR-10 and STL-10 datasets with standard CNN architecture. Experimental setup is the same as \cite{miyato-iclr-2018}, and FID is the score for the comparison. Results are presented in the  third to fifth rows of Table \ref{fid_score}. In addition to settings reported using synthetic dataset, we have additional settings and ablation study for image datasets, which are reported in Section 5 of Supplementary Material. The results confirm the stability of our model, and our method outperforms SN-GAN on the CIFAR-10 dataset. Interestingly, when we replace ``log" by ``hinge loss" functions in the discriminator as in \cite{miyato-iclr-2018}, our ``hinge loss" version performs even better with FID = 22.95, compared to FID = 25.5 of SN-GAN. It is worth noting that our model is trained with the default parameters $\lambda_p = 1.0$ and $\lambda_r = 1.0$. Our generator requires about 200K iterations with the mini-batch size of 64. When we apply our ``hinge loss" version on STL-10 dataset similar to \cite{miyato-iclr-2018}, our model can achieve the FID score 36.19 for this dataset, which is also better than SN-GAN (FID = 43.2). We also compare our model to SN-GAN on CIFAR-10 with similar ResNet architecture as in \cite{miyato-iclr-2018}. Our FID = 17.61 $\pm$ .30 significantly outperforms that of SN-GAN with FID = 21.70 $\pm$ .21. It again confirms the robustness and quality of our proposed GAN model.

\section{Conclusion}

We propose a robust AE-based GAN model with novel distance constraints, called Dist-GAN, that can address the mode collapse and gradient vanishing effectively. Our model is different from previous work: (i) We propose a new generator objective using ``discriminator-score distance''. (ii) We propose to couple the convergence of the discriminator with that of the AE by considering reconstructed samples as ``real'' samples. (iii) We propose to  regularize AE by ``latent-data distance constraint'' in order to  prevent the generator from falling into mode collapse settings. Extensive experiments demonstrate that our method can approximate multi-modal distributions. Our method reduces drastically the mode collapse for MNIST-1K. Our model is stable and does not suffer from mode collapse for MNIST, CelebA, CIFAR-10 and STL-10 datasets. Furthermore, we achieve better FID scores than previous works. These demonstrate the effectiveness of the proposed Dist-GAN. Future work applies our proposed Dist-GAN to different computer vision tasks \cite{hoang-acmm-2017,yiluan-cvpr-2018}.

\section*{Acknowledgement}

This work was supported by both ST Electronics and the National Research Foundation(NRF), Prime Minister's Office, Singapore under Corporate Laboratory @ University Scheme (Programme Title: STEE Infosec - SUTD Corporate Laboratory).

\clearpage

%
%
%

\bibliographystyle{splncs04}
\bibliography{../../../../biblio/biblio}

\begin{thebibliography}{10}
\providecommand{\url}[1]{\texttt{#1}}
\providecommand{\urlprefix}{URL }
\providecommand{\doi}[1]{https://doi.org/#1}

\bibitem{gan_source}
\url{https://github.com/LynnHo/DCGAN-LSGAN-WGAN-WGAN-GP-Tensorflow}

\bibitem{gan_source_b}
\url{https://github.com/carpedm20/DCGAN-tensorflow}

\bibitem{arjovsky-arxiv-2017a}
Arjovsky, M., Bottou, L.: Towards principled methods for training generative
  adversarial networks. arXiv preprint arXiv:1701.04862  (2017)

\bibitem{arjovsky-arxiv-2017}
Arjovsky, M., Chintala, S., Bottou, L.: Wasserstein generative adversarial
  networks. ICML  (2017)

\bibitem{berthelot-arxiv-2017}
Berthelot, D., Schumm, T., Metz, L.: Began: Boundary equilibrium generative
  adversarial networks. arXiv preprint arXiv:1703.10717  (2017)

\bibitem{burda-arxiv-2015}
Burda, Y., Grosse, R., Salakhutdinov, R.: Importance weighted autoencoders.
  arXiv preprint arXiv:1509.00519  (2015)

\bibitem{che-arxiv-2016}
Che, T., Li, Y., Jacob, A.P., Bengio, Y., Li, W.: Mode regularized generative
  adversarial networks. CoRR  (2016)

\bibitem{chen-arxiv-2016}
Chen, X., Duan, Y., Houthooft, R., Schulman, J., Sutskever, I., Abbeel, P.:
  Infogan: Interpretable representation learning by information maximizing
  generative adversarial nets. In: Advances in Neural Information Processing
  Systems. pp. 2172--2180 (2016)

\bibitem{donahue-arxiv-2016}
Donahue, J., Kr{\"a}henb{\"u}hl, P., Darrell, T.: Adversarial feature learning.
  arXiv preprint arXiv:1605.09782  (2016)

\bibitem{dumoulin-arxiv-2016}
Dumoulin, V., Belghazi, I., Poole, B., Lamb, A., Arjovsky, M., Mastropietro,
  O., Courville, A.: Adversarially learned inference. arXiv preprint
  arXiv:1606.00704  (2016)

\bibitem{goodfellow-nips-2016}
Goodfellow, I.: Nips 2016 tutorial: Generative adversarial networks. arXiv
  preprint arXiv:1701.00160  (2016)

\bibitem{goodfellow-nisp-2014}
Goodfellow, I., Pouget-Abadie, J., Mirza, M., Xu, B., Warde-Farley, D., Ozair,
  S., Courville, A., Bengio, Y.: Generative adversarial nets. In: NIPS. pp.
  2672--2680 (2014)

\bibitem{gulrajani-arxiv-2017}
Gulrajani, I., Ahmed, F., Arjovsky, M., Dumoulin, V., Courville, A.C.: Improved
  training of wasserstein gans. In: Advances in Neural Information Processing
  Systems. pp. 5767--5777 (2017)

\bibitem{yiluan-cvpr-2018}
Guo, Y., Cheung, N.M.: Efficient and deep person re-identification using
  multi-level similarity. In: CVPR (2012)

\bibitem{heusel-arxiv-2017}
Heusel, M., Ramsauer, H., Unterthiner, T., Nessler, B., Hochreiter, S.: Gans
  trained by a two time-scale update rule converge to a local nash equilibrium.
  In: Advances in Neural Information Processing Systems. pp. 6626--6637 (2017)

\bibitem{hoang-acmm-2017}
Hoang, T., Do, T.T., Le~Tan, D.K., Cheung, N.M.: Selective deep convolutional
  features for image retrieval. In: Proceedings of the 2017 ACM on Multimedia
  Conference. pp. 1600--1608. ACM (2017)

\bibitem{kingma-arxiv-2013}
Kingma, D.P., Welling, M.: Auto-encoding variational bayes. arXiv preprint
  arXiv:1312.6114  (2013)

\bibitem{larsen-arxiv-2015}
Larsen, A.B.L., S{\o}nderby, S.K., Larochelle, H., Winther, O.: Autoencoding
  beyond pixels using a learned similarity metric. arXiv preprint
  arXiv:1512.09300  (2015)

\bibitem{li-arxiv-2017}
Li, C., Alvarez-Melis, D., Xu, K., Jegelka, S., Sra, S.: Distributional
  adversarial networks. arXiv preprint arXiv:1706.09549  (2017)

\bibitem{lucic-2017-arxiv}
Lucic, M., Kurach, K., Michalski, M., Gelly, S., Bousquet, O.: Are gans created
  equal? a large-scale study. CoRR  (2017)

\bibitem{makhzani-arxiv-2015}
Makhzani, A., Shlens, J., Jaitly, N., Goodfellow, I.: Adversarial autoencoders.
  In: International Conference on Learning Representations (2016)

\bibitem{metz-arxiv-2016}
Metz, L., Poole, B., Pfau, D., Sohl-Dickstein, J.: Unrolled generative
  adversarial networks. ICLR  (2017)

\bibitem{miyato-iclr-2018}
Miyato, T., Kataoka, T., Koyama, M., Yoshida, Y.: Spectral normalization for
  generative adversarial networks. ICLR  (2018)

\bibitem{radford-arxiv-2015}
Radford, A., Metz, L., Chintala, S.: Unsupervised representation learning with
  deep convolutional generative adversarial networks. arXiv preprint
  arXiv:1511.06434  (2015)

\bibitem{rezende-icml-2014}
Rezende, D.J., Mohamed, S., Wierstra, D.: Stochastic backpropagation and
  approximate inference in deep generative models. In: ICML. pp. 1278--1286
  (2014)

\bibitem{salimans-nisp-2016}
Salimans, T., Goodfellow, I., Zaremba, W., Cheung, V., Radford, A., Chen, X.:
  Improved techniques for training gans. In: NIPS. pp. 2234--2242 (2016)

\bibitem{warde-arxiv-2016}
Warde-Farley, D., Bengio, Y.: Improving generative adversarial networks with
  denoising feature matching. ICLR  (2017)

\bibitem{wu-arxiv-2016}
Wu, Y., Burda, Y., Salakhutdinov, R., Grosse, R.: On the quantitative analysis
  of decoder-based generative models. ICLR  (2017)

\bibitem{yazici-arxiv-2018}
Yaz{\i}c{\i}, Y., Foo, C.S., Winkler, S., Yap, K.H., Piliouras, G.,
  Chandrasekhar, V.: The unusual effectiveness of averaging in gan training.
  arXiv preprint arXiv:1806.04498  (2018)

\bibitem{zhao-arxiv-2016}
Zhao, J., Mathieu, M., LeCun, Y.: Energy-based generative adversarial network.
  ICLR  (2017)

\end{thebibliography}

\section*{Supplementary Material}

\subsection*{1. 1D demo}

We implement 1D demos of GAN, WGAN-GP, MDGAN, VAEGAN, and our Dist-GAN to illustrate the behavior of each method (See links of demo videos). In the implementation, we keep the same all parameters and network size (one input layer, one hidden layer with size = 4, and one output layer) for all methods. We observe that baseline GAN, MDGAN cannot approximate the data distribution and suffer from the gradient vanishing. WGAN-GP can approximate data distribution at some points, but diverges later. Moreover, WGAN-GP is sensitive to the penalty values as we show its results with different penalty values (0.1, 0.5, and 1.0). Using a  smaller value may not sufficient to provide strong gradient so that the method can approximate real distribution. On the other hand, using large value may cause divergence. VAEGAN diverges from real data distribution and is slightly collapsed at the end. Our method converges to the data distribution and does not suffer from other issues. Note that in the demo videos, the discriminator's learning curve is blue, and the generator's one is green.

\begin{itemize}
\item GAN: https://www.youtube.com/watch?v=eisFNXbGaNI
\item MDGAN: https://www.youtube.com/watch?v=5FSqZle6Ot8
\item VAEGAN: https://www.youtube.com/watch?v=587z8VBcvvQ
\item WGAN-GP: https://www.youtube.com/watch?v=5MDBwdfD5rY
\item Dist-GAN: https://www.youtube.com/watch?v=IjbdMNo4m\_8
\end{itemize}

%\begin{figure*}
%\centering
%\caption{Screenshots of demos on 1D data of our GAAN, GAN, WGAN-GP, MDGAN and VAEGAN. See the animations in attached videos. Learning curves: Blue line is discriminator loss, and green line is the generator loss.}
%\label{toy_id_demo}
%\end{figure*}

\subsection*{2. 2D synthetic data}

\begin{table}
\centering
\caption{Network structures for synthetic data experiment.}
\begin{tabular}{c | c c c c}
           & $d_\mathrm{in}$ & $d_\mathrm{out}$ & $N_\mathrm{h}$ & $d_\mathrm{h}$ \\ 
\hline
\hline
Encoder ($E$)            & 2     & 2      & 3        & 128 \\
\hline
Generator ($G$)        & 2     & 2      & 3        & 128 \\
\hline
Discriminator ($D$)  & 2     & 1      & 3        & 128 \\
\end{tabular}
\label{toy_network}
\end{table}

\begin{figure*}
\centering
\includegraphics[scale=0.5]{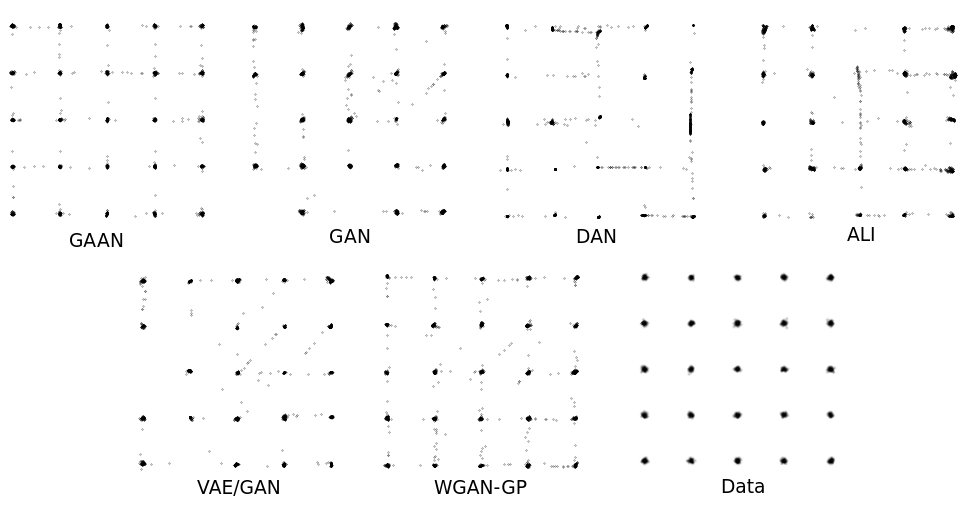}
\caption{Visualizing samples generated by different methods on the synthetic dataset.}
\label{visual_toy_mode}
\end{figure*}

This section describes our model architecture (Table \ref{toy_network}), additional quantitative results (Table \ref{tbl_toydata_comparison_01}) and additional visualization of generated samples (Fig. \ref{visual_toy_mode}) from different methods. 

\begin{table*}
\centering
\footnotesize
\caption{Synthetic data results. Columns indicate the number of covered modes, and the number of registered samples among 2000 generated samples, and two types of Total Variation (TV). We evaluate three versions of Dist-GAN$_1$ regarding reconstructed samples as ``fake", ``none" or ``real" samples.}
\begin{tabular}{c | c | c | c | c}
Method & \#registered modes & \#registered points & TV (True) & TV (Differential) \\ 
\hline
\hline
GAN \cite{goodfellow-nisp-2014}     & 21.50 $\pm$ 2.00     & 1875.63 $\pm$ 42.53  & 1.00 $\pm$ 0.00 & 0.94 $\pm$ 0.02  \\
\hline
DAN-2S \cite{li-arxiv-2017}         & 14.07 $\pm$ 2.85     & 986.33 $\pm$ 160.80  & 0.99 $\pm$ 0.02 & 0.70 $\pm$ 0.09 \\
\hline
VAE-GAN \cite{larsen-arxiv-2015} & 21.13 $\pm$ 2.33 & 1816.19 $\pm$ 61.23 & 0.98 $\pm$ 0.15 & 0.93 $\pm$ 0.21   \\
\hline
ALI \cite{dumoulin-arxiv-2016}      & 22.00 $\pm$ 2.63     & 1490.25 $\pm$ 194.38  & 0.93 $\pm$ 0.11 & 0.68 $\pm$ 0.15  \\
\hline
WGAN-GP \cite{gulrajani-arxiv-2017}   & 24.04 $\pm$ 1.16    & 1766.25 $\pm$ 79.48   & 0.54 $\pm$ 0.35 & 0.43 $\pm$ 0.31  \\
\hline
%GAAN$_1$        & 9     & 1353    \\
%\hline
%GAAN$_1$        & -     & 1602  &  \\
GAN$_1$        & 23.75 $\pm$ 0.71 & 1899.62 $\pm$ 26.48 & 0.90 $\pm$ 0.29 & 0.86 $\pm$ 0.27 \\
\hline
GAN$_2$        & 24.50 $\pm$ 1.07 & 1809.62 $\pm$ 113.54 & 0.42 $\pm$ 0.32 & 0.33 $\pm$ 0.27 \\
\hline
%GAAN$_1$        & 24.52 $\pm$ 1.00 & 1652.89 $\pm$ 148.22 & 0.47 $\pm$ 0.26 & 0.31 $\pm$ 0.19 \\
%\hline
Dist-GAN$_1$ (fake)        &  17.67 $\pm$ 1.89 &  1678.50 $\pm$  52.81 &  1.25 $\pm$ 0.32 & 1.31 $\pm$  0.44 \\
\hline
Dist-GAN$_1$ (none)        &  21.96 $\pm$ 1.96 &  1710.62 $\pm$ 51.65 &  1.04 $\pm$ 0.26 &  0.99 $\pm$ 0.35 \\
\hline
Dist-GAN$_1$ (real)        &  23.62 $\pm$ 1.05 &  1756.12 $\pm$ 31.69 &  0.87 $\pm$ 0.17 &  0.80 $\pm$ 0.22 \\
\hline
Dist-GAN$_2$        & 25.00 $\pm$ 0.00 & 1719.25 $\pm$ 49.77 & 0.31 $\pm$ 0.09 & 0.18 $\pm$ 0.06 \\
\hline
Dist-GAN       & 25.00 $\pm$ 0.00 & 1827.38 $\pm$ 25.60 & 0.20 $\pm$ 0.03 & 0.12 $\pm$ 0.02\\
\end{tabular}
\label{tbl_toydata_comparison_01}
\end{table*}

\subsection*{3. CelebA dataset}

\begin{figure*}
\centering
\includegraphics[scale=0.3]{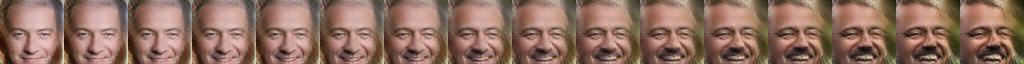}
\includegraphics[scale=0.3]{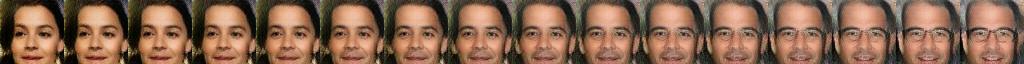}
\includegraphics[scale=0.3]{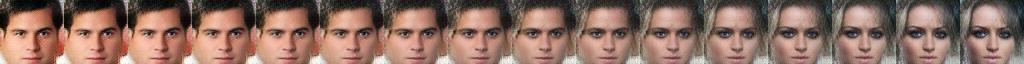}
\includegraphics[scale=0.3]{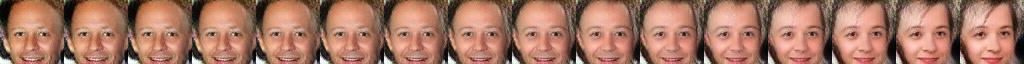}
\includegraphics[scale=0.3]{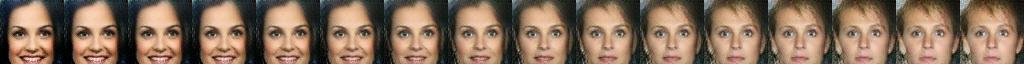}
\caption{Samples generated by our method from interpolated latents.}
\label{interpolation}
\end{figure*}

Here we  demonstrate that our model does not have  over-fitting.
Also, our model allows control using latent space samples. We interpolate the latent variables to generate faces as shown in Fig. \ref{interpolation}. The transitions in the generated faces are smooth, and interpolated faces are realistic, suggesting that our model can be used  to generate smooth faces from latents.

\subsection*{4. CIFAR-10 and STL-10 datasets}

\begin{figure}
\centering
\includegraphics[scale=0.51]{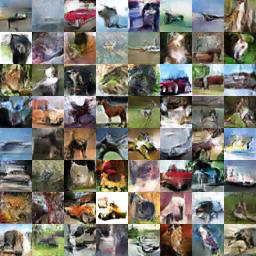}
\includegraphics[scale=0.51]{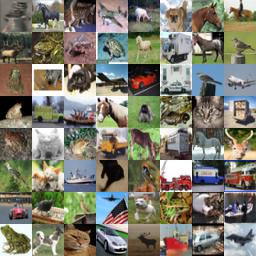}
\includegraphics[scale=0.34]{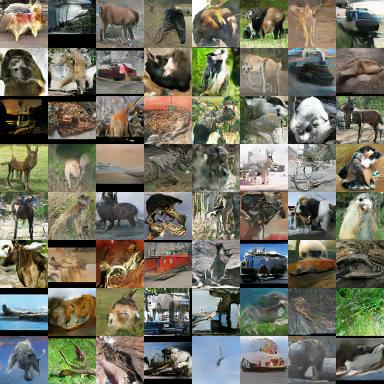}
\includegraphics[scale=0.34]{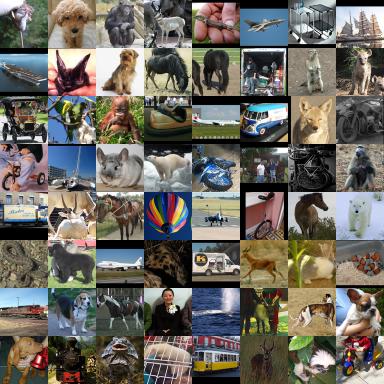}
\caption{Generated samples using our method trained on CIFAR-10 (first row) and STL-10 (second row). Generated samples are on the left and real samples on the right.}
\label{cifar_generated_samples}
\end{figure}

We visualize generated samples on CIFAR-10 and STL-10 datasets, Fig. \ref{cifar_generated_samples}. The output images are shape, and have sufficient variation. This is also shown with FID score in the main paper.

\begin{figure}
%Put here to reduce too much white space after table
\centering
\scriptsize
\includegraphics[scale=0.5]{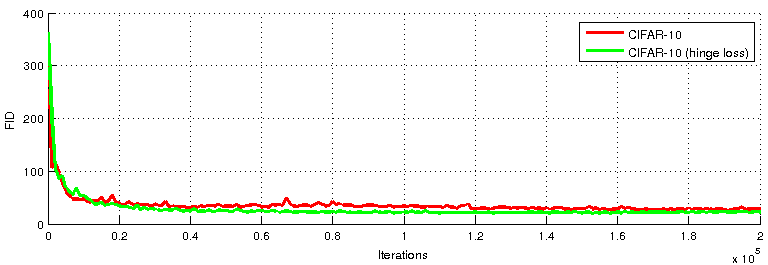}
\vspace*{-4mm}%Put here to reduce too much white space after table
\caption{The convergence of FID scores on CIFAR-10 dataset.}
\label{fid_curve}
\end{figure}

We found that, in each iteration, training the generator twice can improve convergence and does not cause instability to our model. In particular, first, we train the generator as part of AE for realistic samples reconstruction. Then, we train the generator to confuse the discriminator in order to generate better images. The AE may converge to some local minimum, but training the generator again could direct the AE to converge  to some saddle point, where the generator can generate quality images. The convergence of FID scores is shown in Fig. \ref{fid_curve} when training our model on CIFAR-10.

\subsection*{5. Ablation study on CIFAR-10}

\begin{table}
\centering
\scriptsize
\caption{Different configurations in our ablation study. $\mathcal{L}_G, L_C, L_W, L_P$ indicates the use of our proposed generator objective (discriminator score distance), reconstruction constraints, regularization of AE (latent-data distance) and gradient penalty respectively. $\checkmark$: use, $\times$: no use. `*': collapsed.}
\begin{tabular}{c | c | c | c | c | c | c | c }
\hline
 Use & \textbf{Dist-GAN$_1$} & \textbf{Dist-GAN$_2$} & \textbf{Dist-GAN$_3$} & \textbf{Dist-GAN$_4$} & \textbf{Dist-GAN$_5$} & \textbf{Dist-GAN$_6$} & \textbf{Dist-GAN} \\
\hline
$\mathcal{L}_G$ & $\checkmark$ & $\checkmark$ & $\times$ & $\checkmark$ & $\checkmark$ & $\checkmark$ & $\checkmark$ \\
$L_C$ & $\checkmark$ & $\checkmark$ & $\checkmark$ & $\times$ & $\checkmark$ & $\times$ & $\checkmark$ \\
$L_W$ & $\times$ & $\times$ & $\checkmark$ & $\checkmark$ & $\checkmark$ & $\checkmark$ & $\checkmark$ \\
$L_P$ & $\times$ & $\checkmark$ & $\checkmark$ & $\checkmark$ & $\times$ & $\times$ & $\checkmark$
\end{tabular}
\label{setting_ablation}
\end{table}

This ablation study complements the synthetic data experiments. We remove each of the proposed components $\mathcal{L}_G, L_C, L_W, L_P$ (denote the generator objective, the reconstruction constraints of discriminator, the AE regularization and the gradient penalty, resp.) from our model and then evaluate on CIFAR-10 dataset. Ablation results (Table \ref{fid_score_ablation}) suggests that the generator objective (discriminator score distance) can significantly reduce mode collapse (see Dist-GAN$_3$). Also,  regularization of Autoencoder (latent-data distance) can reduce mode collapse (see Dist-GAN$_2$).

%, and two other components: the gradient penalty (Dist-GAN$_5$) and reconstruction constraints (Dist-GAN$_4$) have lower impacts on the dataset. In this experiment, we again see that using reconstruction constraint (Dist-GAN$_5$) improves the FID score from (Dist-GAN$_6$) without using it.

\begin{table}
\centering
\scriptsize
\caption{Ablation study on CIFAR dataset with CNN architecture.}
\begin{tabular}{c | c | c | c | c | c | c }
\hline
 & \textbf{Dist-GAN} & \textbf{Dist-GAN$_2$}  & \textbf{Dist-GAN$_3$} & \textbf{Dist-GAN$_4$} & \textbf{Dist-GAN$_5$} & \textbf{Dist-GAN$_6$} \\
\hline
\textbf{CIFAR} & \textbf{28.23} & 36.09 & $>$200* & 29.80 & 28.50 & 30.37 \\
\end{tabular}
\label{fid_score_ablation}
\end{table}

\end{document}